\documentclass{article}

\usepackage{iftex}
\ifPDFTeX
\usepackage[T1]{fontenc}
\fi
\usepackage{graphicx}
\usepackage{booktabs}
\usepackage{array}
\usepackage{amsmath,amssymb}
\usepackage{hyperref}
\usepackage{etoolbox}

\usepackage[preprint]{icml2026}

\ifPDFTeX\else
\usepackage{fontspec}
\fi
\newcommand{\authorcontact}[2]{%
  \begin{minipage}[t]{0.32\textwidth}
    \centering
    {\normalsize\bfseries #1\par}
    \vspace{2pt}
    {\footnotesize\ifstrempty{#2}{\strut}{\href{mailto:#2}{#2}}\par}
  \end{minipage}%
}

\makeatletter
\newcommand{\printPreprintNotice}{%
  \global\icml@noticeprintedtrue
  {\let\thefootnote\relax
    \footnotetext{\hspace*{-\footnotesep}\Notice@String}}%
}
\makeatother

\icmltitlerunning{Multimodal Models Can See, But Can They Think?}
\icmlkeywords{Visual Reasoning, Large Language Models, Vision-Language Models, Fine-Tuning}

\begin{document}

\twocolumn[
\icmltitle{Multimodal Models Can See, But Can They Think? A Study of Turkish Visual Reasoning}

\begin{center}
\authorcontact{Berke Bunyamin Sule}{bunyamin.sule@std.yildiz.edu.tr}\hfill
\authorcontact{Yavuz Selim Aygan}{selim.aygan@std.yildiz.edu.tr}\hfill
\authorcontact{Ahmed Zeer}{ahmed.zeer@std.yildiz.edu.tr}
\par\vspace{10pt}
\authorcontact{Elif Ince}{elif.ince@std.yildiz.edu.tr}\hfill
\authorcontact{Ali Mert Temizsoy}{mert.temizsoy@std.yildiz.edu.tr}\hfill
\authorcontact{Ozgur Agar}{ozgur.agar@std.yildiz.edu.tr}
\par\vspace{10pt}
\authorcontact{Himmet Toprak Kesgin}{tkesgin@yildiz.edu.tr}\hfill
\authorcontact{Muzaffer Kaan Yuce}{kaan.yuce@yildiz.edu.tr}\hfill
\authorcontact{Mehmet Fatih Amasyali}{amasyali@yildiz.edu.tr}
\par\vspace{12pt}
{\small\itshape Department of Computer Engineering, Yildiz Technical University\par
Istanbul, Turkiye\par}
\end{center}
\vskip 0.2in
]

\printPreprintNotice
\raggedbottom
\hypersetup{
  pdftitle={Multimodal Models Can See, But Can They Think? A Study of Turkish Visual Reasoning},
  pdfsubject={Preprint on Turkish visual reasoning in vision-language models},
  pdfauthor={Berke Bunyamin Sule, Yavuz Selim Aygan, Ahmed Zeer, Elif Ince, Ali Mert Temizsoy, Ozgur Agar, Himmet Toprak Kesgin, Muzaffer Kaan Yuce, Mehmet Fatih Amasyali}
}

\begin{abstract}
Vision-Language Models have had massive leaps in recent years. Yet, when tested on high-level cognitive tasks like pattern completion, abstract logic, and relational reasoning, they still fall short. This limitation stems from the models' lack of ability to find out underlying rules and maintain constant intermediate steps, rather than merely deciding on the correct option. To research these shortcomings, this study analyzes visual reasoning performance throughout a multilingual evaluation framework encompassing the BİLSEM, MM-IQ, Visual Puzzles, and PuzzleVQA datasets. We evaluate various data mixing strategies and supervised fine-tuning (SFT) methods. We observed that a correct answer does not always mean the reasoning chain is valid. Because of this, we adopt an LLM-as-a-Judge approach to assess the quality of reasoning beyond final answer accuracy. Our experimental results reveal that baseline models often exhibit more stable reasoning compared to fine-tuned counterparts. Most importantly, we observe that SFT can lead to model collapse, manifesting as infinite text loops or empty outputs. The study concludes that current VLMs still struggle with abstract visual reasoning.

\end{abstract}
\section{Introduction}
Vision-Language Models (VLMs) demonstrate high performance in tasks such as object recognition and scene description. However, their performance drops significantly when it comes to problems requiring abstract pattern analysis, matrix completion, and relational logic. This decline is associated with the necessity of discovering rules, maintaining intermediate steps consistently, and explicitly demonstrating the rationale, rather than merely selecting the correct option. Since visual reasoning questions simultaneously trigger different cognitive components like symbolic generalization, relational comparison, and contextual consistency, they generate numerous intermediate stages where the model can make mistakes. Consequently, evaluating only the final answer is insufficient for determining the validity of the model's internal reasoning.

In the literature, the areas where VLMs are strong are mostly situated at the level of "visual perception/superficial matching". Conversely, in questions requiring rule discovery, the model must both accurately extract visual cues and establish a consistent relationship between them. For instance, a pattern in a matrix might have a changing rule at both the row and column levels; in this case, the model is expected to evaluate multiple hypotheses on a single image and consistently maintain its rationale. Similarly, in visual puzzles, the "correct answer" often depends on a relational transformation or a hidden rule rather than a single object. Such tasks test not only the model's visual recognition ability but also its capacity for systematic generalization.

This study aims to systematically investigate where VLMs struggle with these components in visual reasoning tasks. Instead of merely conducting a performance comparison, it also analyzes which training strategies produce more consistent reasoning and in which situations the model resorts to "shortcut" behaviors. Thus, an evaluation baseline has been established that characterizes model behaviors in greater detail, bringing together both quantitative and qualitative indicators.

The contributions of the study are gathered along three axes:
\begin{itemize}
    \item A multi-lingual and multi-source evaluation framework consisting of 5,240 samples from the BİLSEM (806), MM-IQ (2,268), Visual Puzzles (466), and PuzzleVQA (1,700) datasets is introduced. This framework increases linguistic diversity by including both Turkish and English questions, enabling the quantitative measurement of visual reasoning capabilities and the observation of how models behave in different language contexts.
    \item The effects of different data mixing strategies and Supervised Fine-Tuning (SFT) processes on performance, reasoning consistency, and the risk of model collapse are systematically analyzed. Specifically, through the proposed SFT strategy using the PuzzleVQA+MMIQ (3pro) data mixture, the 24.7\% baseline accuracy is improved to 48.8\%. This nearly two-fold performance increase empirically demonstrates that models can genuinely learn logical deduction, while the broader analysis answers not only "which data yields better results?" but also "which data mixtures can disrupt the model's consistency?".
    \item An LLM-as-a-Judge evaluation methodology is adopted to assess the structural quality of the reasoning sequence. This method enables the explicit identification of cases where models produce correct answers despite flawed reasoning trajectories, providing a deeper understanding of internal logic beyond conventional final answer accuracy.
\end{itemize}
In the remainder of the paper, the related literature is first summarized, followed by an explanation of the datasets and data processing steps. This is followed by model training, evaluation methodology, and experimental results; in the final section, the findings are discussed, and suggestions for future work are presented.

\begin{figure}[t] 
\centering
\begin{minipage}{\linewidth} 
    \centering
    \includegraphics[width=\linewidth]{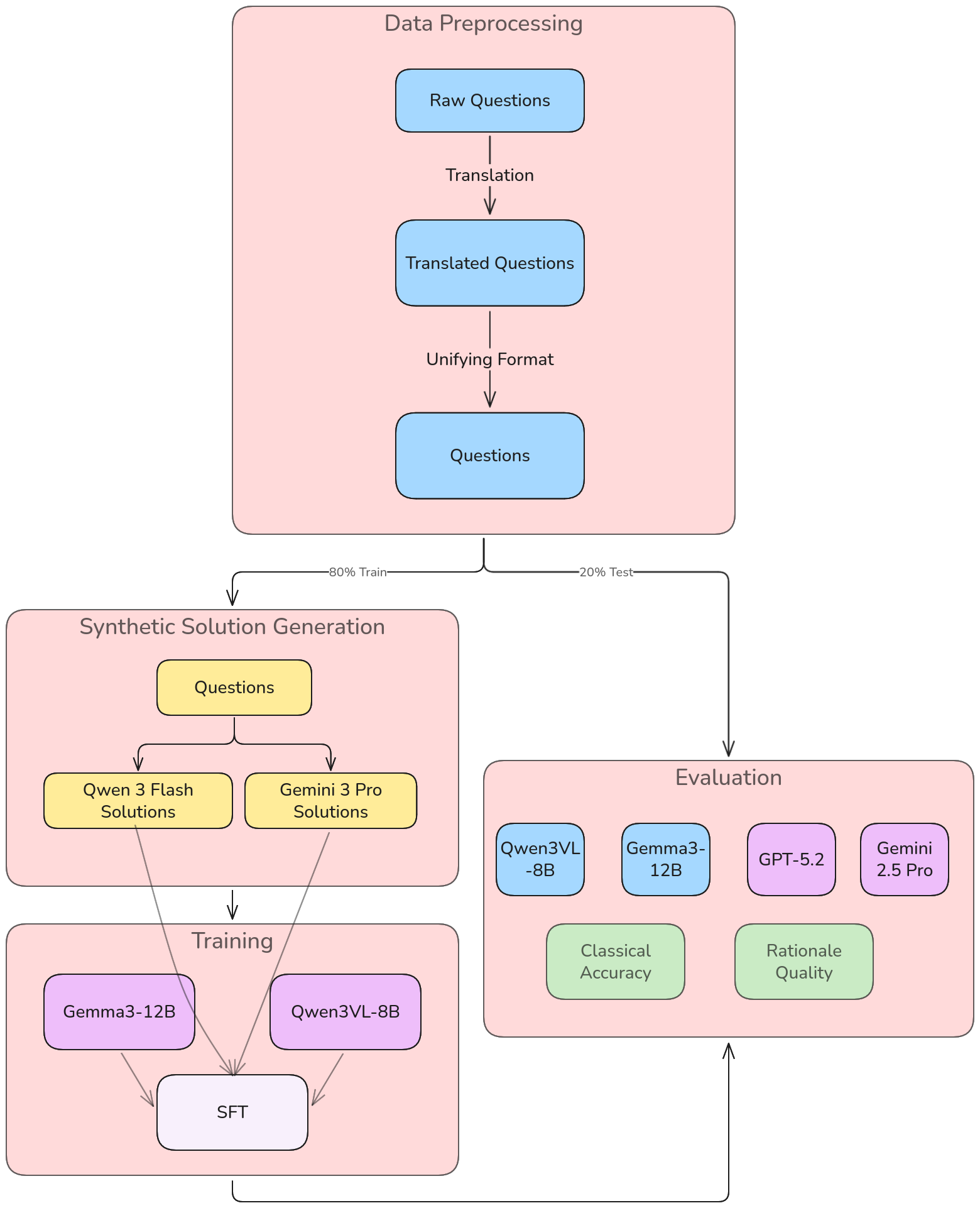} 
    \vspace{4pt}
    \small
\end{minipage}
\caption{An overview of the proposed evaluation and training framework.}
\label{fig:methodology-diagram}
\end{figure}

\section{Related Work}
This section summarizes related studies supporting the four main axes of this research: visual reasoning benchmarks, fine-tuning strategies and model collapse, VLM training with reinforcement learning (RL), and evaluation methodology.

\subsection{Abstract Visual Reasoning Benchmarks}
Abstract Visual Reasoning (AVR) encompasses evaluation tasks designed to minimize the impact of language-based shortcuts. The primary inspiration in this domain is the Raven's Progressive Matrices (RPM) test; \citet{ref_zhang} adapted this psychology-rooted paradigm for computer vision and created the RAVEN dataset, which evaluates how rules applied to an object change through pattern matrices. 

Studies applying this framework to the evaluation of VLMs have yielded striking findings regarding the extent to which models succeed in rule discovery. MARVEL, developed by \citet{ref_jiang}, includes 770 puzzles consisting of six basic information patterns, geometric and abstract shapes, and five different task configurations. The study paired each abstract reasoning question with complementary perception questions to distinguish whether the model's accuracy stemmed from genuine perception and reasoning capacity or simply from guessing strategies. 

Focusing on this issue from the perspective of matrix reasoning, MaRs-VQA ~\citep{ref_cao} systematically measures the performance of VLMs on RPM-like abstract visual questions. An important finding that intersects with our study is the report that SFT conducted on MaRs-VQA does not degrade the model's general VLM capabilities—meaning catastrophic forgetting does not occur; this is an optimistic result showing that domain-focused fine-tuning is possible.

MM-IQ, developed by \citet{ref_cai}, provides a detailed evaluation framework aiming to disentangle linguistic proficiency and domain knowledge from the measurement of abstract reasoning capacity. Containing 2,710 test questions across eight different reasoning paradigms, such as temporal motion, spatial relationships, logical operations, and 2D/3D geometry, this dataset showed that even state-of-the-art models perform near random chance levels. This result strikingly illustrates that current VLMs have not yet reached basic human reasoning capacities.

PuzzleVQA~\citep{ref_chia}, which focuses on abstract questions of the visual puzzle type, reveals that models exhibit significant performance disparities depending on the task type. The experimental findings in our study confirm this disparity: models achieve much higher success in visual puzzle-type tasks compared to IQ-type abstract questions; this suggests that visual pattern recognition and abstract rule discovery capacities are independent of each other.

\subsection{Fine-Tuning Strategies, Model Collapse, and RL vs. SFT}
Post-training strategies aimed at enhancing visual reasoning capacity constitute one of the most active research areas recently. A comparative study by \citet{ref_chu} concretely showed the differing effects of SFT and RL on generalization capacity: SFT memorizes training data and fails to generalize in out-of-distribution (OOD) scenarios; whereas RL, especially when trained with outcome-based rewards, can generalize to both rule-based and visual variants. This "SFT memorizes, RL generalizes" thesis aligns with one of the primary findings of our study: the baseline model outperforming SFT-applied models in certain configurations. Conversely, \citet{ref_yu} present a more nuanced picture: the effectiveness of SFT is conditional on factors such as model capacity, data scale, and data distribution; in smaller models and with limited data, SFT can exhibit more reliable reasoning behaviors compared to RL.

The risk of model collapse is a documented, significant threat in training processes involving SFT. Catastrophic forgetting occurs when gradient updates performed for a new task overwrite parameters that are critical for previous tasks~\citep{ref_mccloskey}. In the context of VLM training, this risk is particularly sensitive to hyperparameters like weight initialization, learning rate, and context length. The "thinking loop" (where the model infinitely repeats the same text) and the high rate of empty outputs observed in our study are practical reflections of this theoretical risk. Parameter-efficient fine-tuning methods like LoRA mitigate this risk but do not entirely eliminate it.

GRPO (Group Relative Policy Optimization) has emerged as an algorithm that stabilizes policy updates with verification-based rewards, inspired by DeepSeek-R1's ~\citep{ref_guo} use of reinforcement learning to strengthen LLM reasoning capacity. Vision-R1, developed by \citet{ref_huang}, is an early work that adapted GRPO for VLMs and achieved high success on the MathVista benchmark. However, the effectiveness of GRPO in smaller models remains limited due to its reliance on model capacity~\citep{ref_yu}.

\subsection{Evaluation Methodology}
It is increasingly recognized in the literature that the classical multiple-choice accuracy metric is insufficient for evaluating visual reasoning tasks. \citet{ref_jiang} revealed that arriving at the correct answer and executing correct reasoning can be independent of each other by reporting instances where the model produced the right answer but failed the perception question. Driven by this motivation, the LLM-as-a-Judge approach was developed to measure rationale quality~\citep{ref_zheng}. In this paradigm, a strong LLM is utilized as an evaluator in open-ended assessment tasks where rule-based or lexical metrics fall short; thereby, the scalability and consistency of human evaluation are combined with the speed of rule-based metrics. Building on this framework, our study employs LLM-based scoring across three dimensions: logical consistency, conciseness and clarity, and alignment with the reference solution.

\section{Dataset and Data Engineering}

\subsection{Datasets Used}
In selecting the datasets, which are the outputs of an original academic study, the fundamental criteria adopted were scale, the inclusion of analytical resolutions, and each question possessing a single correct answer. Accordingly, the utilized datasets are characterized as follows:

\begin{itemize}
    \item PuzzleVQA: Consisting of 2,000 samples, this dataset is at an accessible difficulty level with an average human accuracy of 91\%. It includes three-stage correct solution paths: Caption (describing the visual elements), Explanation (detailing the underlying rules or logic), and Deduction (applying these rules to derive the final answer).
    \item MM-IQ: The human accuracy rate remaining at 51\% indicates that this set is cognitively more complex and distinguishing. This difficulty level is the fundamental element enriching the sample diversity in the study.
    \item Visual Puzzles: With a 75\% average success rate, it bridges the easy and difficult extremes, consolidating the overall balance and heterogeneity of the dataset.
    \item BİLSEM Preparation Questions: Approximately 883 original Turkish questions compiled from preparation materials for the Science and Art Centers (BİLSEM) exams, where gifted students are educated in Turkey, were added to the dataset. The fact that the questions are directly in Turkish and in a standard structure contributes to the linguistic and logical diversity of the dataset.
\end{itemize}

\begin{figure*}[t]
\centering

\begin{minipage}{0.32\linewidth}
\centering
\includegraphics[width=\linewidth]{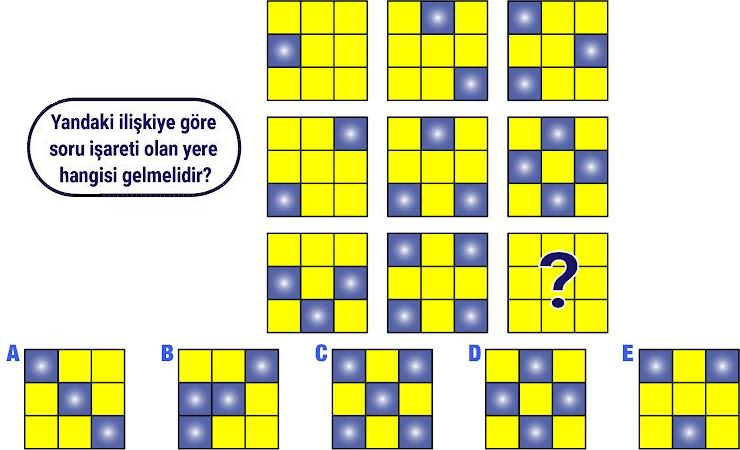}
\small (a) BİLSEM \\ Answer: C
\end{minipage}
\hfill
\begin{minipage}{0.32\linewidth}
\centering
\includegraphics[width=\linewidth]{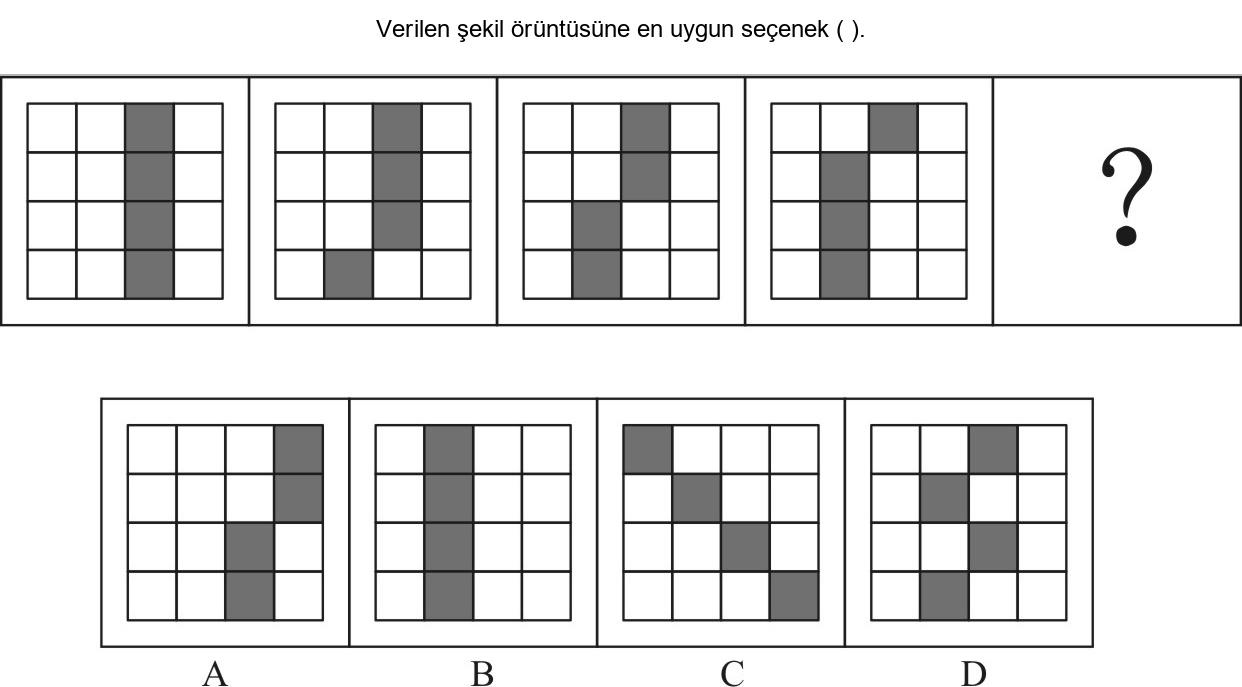}
\small (b) MM-IQ \\ Answer: B
\end{minipage}
\hfill
\begin{minipage}{0.32\linewidth}
\centering
\includegraphics[width=\linewidth]{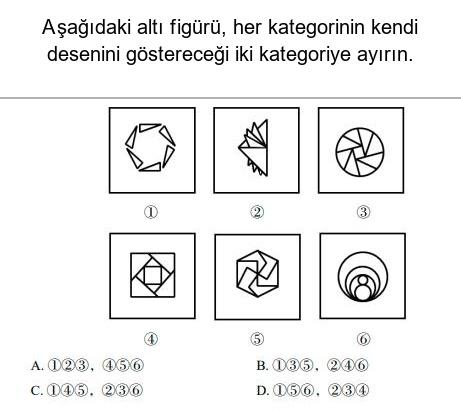}
\small (c) Visual Puzzles \\ Answer: D
\end{minipage}

\caption{Representative samples from different datasets used in this study.}
\label{fig:dataset-samples}
\end{figure*}

\begin{figure}[t]
\centering
\begin{minipage}[c]{0.4\linewidth}
\centering
\includegraphics[width=\linewidth]{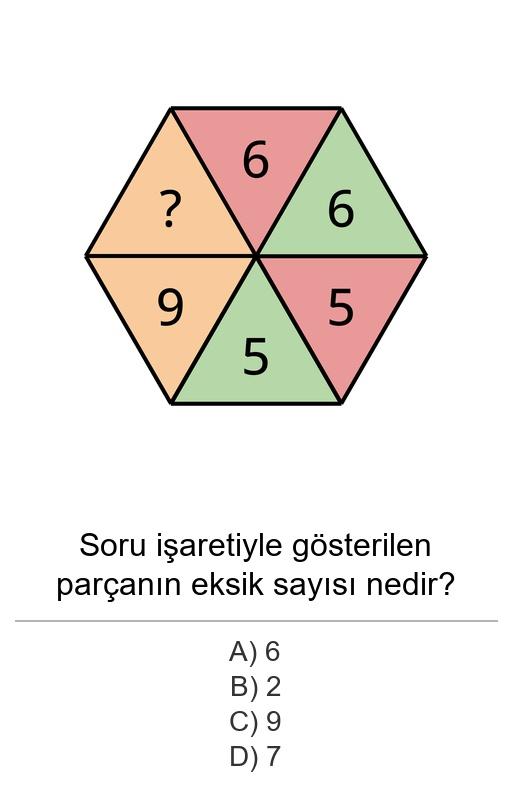}
\par\vspace{6pt}
\small Answer: B) 2
\end{minipage}
\hfill
\begin{minipage}[c]{0.58\linewidth}

\small
\textbf{Caption:} Altı parçaya bölünmüş bir altıgen vardır; renkleri ['kırmızı', 'turuncu', 'turuncu', 'yeşil', 'kırmızı', 'yeşil'] olan ve saat yönünün tersine bir sırayla dizilmiş. Kısımlar sırasıyla [6, '?', 9, 5, 5, 6] sayıları ile gösterilmiştir.

\vspace{6pt}

\textbf{Explanation:} Yeşil kısımlardaki sayıların toplamı 11'e eşittir. Benzer şekilde, turuncu kısımlardaki sayıların toplamı da 11'e eşittir. Dolayısıyla, aynı renkteki kısımlardaki sayıların toplamı 11'dir.

\vspace{6pt}

\textbf{Deduction:} Aynı renkteki parçaların içindeki sayıların toplamı 11 olduğuna göre, turuncu parçadaki eksik sayı 2 olmalıdır.

\end{minipage}

\caption{A sample from the PuzzleVQA dataset with three-stage reasoning annotation (caption, explanation, deduction).}
\label{fig:pvqa-sample}
\end{figure}

Figures~\ref{fig:dataset-samples} and~\ref{fig:pvqa-sample} present representative examples from the datasets used in this study, highlighting their diversity in structure, difficulty, and reasoning requirements. Notably, the original PuzzleVQA dataset inherently includes step-by-step solutions, while detailed solutions for the BİLSEM dataset were manually generated. Thus, both datasets provide complete sets of questions, answer options, and solution paths. In contrast, the MM-IQ and Visual Puzzles datasets contain only questions and answer options, without step-by-step solutions. Finally, to avoid difficulties during the evaluation phase, samples lacking options in the Visual Puzzles dataset were filtered out. To maintain proportional representation across all diverse data sources, each individual dataset was independently split using an 80-20 ratio for training and testing, respectively. From this combined test pool, a unified evaluation subset of 400 samples was selected to conduct the benchmark experiments.


\subsection{Data Processing Pipelines}
Raw data underwent two fundamental processes for efficient processing by the models. First, all textual content in the English-sourced datasets (MM-IQ, Visual Puzzles, PuzzleVQA) was translated into Turkish using Large Language Models while preserving the context. 

In the second stage, an OpenCV-based image merging pipeline was developed to combine the question text, image, and options—which were present as separate components in the source data—into a single image. This system produces a standard and model-friendly single image with a vertical layout: the Turkish question text at the top, the relevant image in the middle, and the aligned options at the bottom. 

\subsection{Synthetic Solution Generation}
The fact that all datasets contain the correct answer helped in determining the path leading to that answer. When choosing among closed-source models, speed and cost were considered. The batch processing capability of the Gemini 3 Pro \citep{ref_gemini3} model and the compact, fast nature of the Qwen3-VL-Flash \citep{ref_qwen3vl}  model were the main criteria for their selection.

The question text was not separately added to the prompt, since it was already embedded in the input image. Instead, to guide the generation process, the known correct answer was directly included in the system prompt. Solutions suggesting the wrong answer were filtered out, and the impact of this was measured. Also, different correct solutions for a single question were obtained through various prompting methods, and the effect of this was also observed.

\begin{table*}[t]
\centering
\caption{Summary of dataset sources and data splitting}
\label{tab1}
\begin{tabular}{|l|l|l|l|l|}
\hline
Data Source & Features & Train & Test & Total \\
\hline
MM-IQ & Question and Answer & 2,168 & 100 & 2,268 \\
Visual Puzzles & Question and Answer & 366 & 100 & 466 \\
PuzzleVQA & Question, Answer, and Solution & 1,600 & 100 & 1,700 \\
BİLSEM Questions & Question, Answer, and Solution & 706 & 100 & 806 \\
\hline
Total & & 4,840 & 400 & 5,240 \\
\hline
\end{tabular}
\end{table*}

\subsection{Token Cost}
This section presents the approximate token consumption of API-based operations throughout the project and the corresponding cost calculations. For synthetic data generation within the scope of the project, Gemini 3 Pro Preview (Batch) and Qwen3-VL-Flash were utilized.

\paragraph{Calculation Method.} The total number of tokens for each request is defined as follows:
\begin{equation}
T^{(i)}=T_{in}^{(i)}+T_{out}^{(i)}
\end{equation}

Total consumption:
\begin{equation}
T_{total}=\sum_{i=1}^{N}T^{(i)}
\end{equation}
where N represents the total number of initial API requests.

The cost, based on the relevant provider's unit token pricing, is calculated as:
\begin{equation}
C=\frac{T_{in}}{10^{6}}p_{in}+\frac{T_{out}}{10^{6}}p_{out}
\end{equation}

In the event of retries, the total number of calls can be scaled as $N_{tot}=N(1+r)$, where r denotes the average retry rate, and the costs increase approximately by a factor of $(1+r)$.

\paragraph{Qwen3-VL-Flash (Alibaba Cloud, billing records).} Token consumption for Qwen3-VL-Flash was calculated by filtering the Alibaba Cloud Model Studio billing records (specifically the "Text ModelTokens" item, rows labeled qwen3-vl-flash). According to these records, the total input tokens are 4,929,410, the total output tokens are 14,968,788, and the total cost is 6.2337 USD.

\begin{table}[t]
\centering
\caption{Token usage and cost obtained from billing records for Qwen3-VL-Flash (Alibaba Cloud).}
\label{tab2}
\begin{tabular}{|l|l|l|}
\hline
Item & Token & Amount (USD) \\
\hline
Input & 4,929,410 & 0.2462 \\
Output & 14,968,788 & 5.9875 \\
\hline
Total & 19,898,198 & 6.2337 \\
\hline
\end{tabular}
\end{table}

\begin{table}[t]
\centering
\small
\setlength{\tabcolsep}{3pt}
\caption{Approximate token usage and cost for Gemini 3 Pro Preview (Batch).}
\label{tab3}
\begin{tabular}{|l|l|l|l|l|}
\hline
Scenario & Input (M) & Output (M) & Total (M) & Cost (USD) \\
\hline
Low & 2.628 & 2.000 & 4.628 & 14.628 \\
Medium & 4.028 & 3.400 & 7.428 & 24.428 \\
High & 5.428 & 4.800 & 10.228 & 34.228 \\
\hline
\end{tabular}
\end{table}

\paragraph{Gemini 3 Pro Preview (Batch, approximate).} Approximately $N\approx4000$ requests were executed on the Gemini side for synthetic data generation. Questions and options were not provided as text in the prompt; they were embedded within the image. Therefore, the entirety of the text input consists solely of the system prompt ($T_{text,in}=157$ tokens). The remaining portion of the input tokens comes from the image representation ($T_{img}$), and since this value varies depending on the resolution, it was approximated using low, medium, and high scenarios. Based on data inspection, output tokens vary between 500 and 1200; the scenarios assume 500, 850, and 1200, respectively. For batch pricing, $p_{in}=1$ USD/1M tokens and $p_{out}=6$ USD/1M tokens were applied. In the scenarios, it was assumed that for each sample $T_{in}=157+T_{img}$ and $T_{out}\in\{500,850,1200\}$, and total tokens were scaled by $N=4000$.

\textit{Note:} For retries, the values in Table 3 can be scaled by the factor $(1+r)$ using the retry rate $r$ (e.g., for $r=0.1$, costs increase by approximately 10\%).

\section{Model Training}
A series of experimental studies were conducted to observe the effects of the created datasets on the reasoning performance of VLMs. The supervised fine-tuning (SFT) method was utilized for training the models. The low-rank adaptation (LoRA) technique was preferred during training due to its cost-effectiveness and its ability to reduce the risk of catastrophic forgetting. The training sessions were performed on the following two base models: Qwen3VL-8B and Gemma3-12B \citep{ref_gemma3} with the configuration shown in Table~\ref{tab:hyperparameters}.

\begin{table}[t]
\caption{Training Hyperparameters}
\centering
\begin{tabular}{|l|r|}
\hline
\textbf{Parameter} & \textbf{Value} \\
\hline
LoRA Rank & 16 \\
LoRA Alpha & 32 \\
Effective Batch Size & 16 \\
Learning Rate & $5 \times 10^{-5}$ \\
Scheduler & Cosine \\
Epoch & 2 \\
\hline
\end{tabular}
\label{tab:hyperparameters}
\end{table}

\section{Evaluation Methodology}

\subsection{The Issue with Classic Accuracy}

Traditional accuracy metrics in multiple-choice questions solely focus on whether the correct option is selected. However, in visual reasoning problems, a model arriving at the right answer does not invariably imply that it followed a correct reasoning process. In some instances, the model might guess the correct option relying on superficial cues, while in other cases, it might employ the right logic but miss the final answer. Consequently, an evaluation strictly based on a true/false binary obscures the model's rationale quality, the consistency of its intermediate steps, and its capacity for rule discovery.

The primary objective here is not merely to generate the final answer; rather, it is to accurately discover the underlying pattern within the problem, consistently apply the derived rule, and explicitly articulate the rationale. Therefore, the evaluation approach must take into account the solution explanation generated by the model alongside the final result. To address this, we adopt the LLM-as-a-Judge approach described in the following subsection.

\subsection{Rationale Quality Evaluation}

A pattern observed in our study clearly exposes the limitation of classic accuracy: some models construct a fundamentally flawed reasoning chain while still arriving at the correct answer, whereas others establish the correct logic but fail at the final selection. Thus, a separate evaluation approach is required to assess the quality of the rationale.

Since embedding-based similarity methods (e.g., cosine similarity) are inadequate for capturing multi-step logical chains, we follow prior work and adopt the LLM-as-a-Judge approach~\citep{llm_as_a_judge}. In this paradigm, a language model acts as an evaluator, scoring aspects of reasoning that rule-based metrics fail to measure. Initial trials were conducted with Cosmos Gemma 9B\citep{ref_cosmosgemma}, but these yielded inconsistent results; subsequently, the Gemini 2.5 Flash API \citep{ref_gemini25} was utilized, providing more stable outputs.

The evaluation is conducted on a 0 to 100 scale across three primary criteria:
\begin{itemize}
    \item \textbf{Logical Consistency:} whether the reasoning steps are consistent with each other and the visual input.
    \item \textbf{Conciseness and Clarity:} whether the model reaches the conclusion directly, avoiding unnecessary repetition and convoluted paths.
    \item \textbf{Alignment with Reference Solution:} the degree to which the generated rationale matches the reference solution in terms of underlying reasoning steps.
\end{itemize}

To reduce potential bias, solution pairs were presented in an A--B format without specifying which solution belonged to which model. A calibration study was carried out on the PuzzleVQA dataset to assess the behavior of the evaluator, comparing solutions that were close to the reference, incorrect, and entirely identical.


As shown in Table~\ref{tab:tab4}, although Qwen3-VL-Flash achieved 89\% accuracy, it falls 54.15 points behind Gemini 3 Pro in Conciseness and Clarity. This indicates that even when a model finds the correct answer, it may rely on unnecessarily complex or verbose reasoning. This evaluation approach therefore serves as a complementary layer to the classic accuracy metric, enabling a more detailed analysis of model behavior. Table~\ref{tab:tab4} reports 
the average scores across the full PuzzleVQA calibration set, providing a 
broader picture of the systematic reasoning gap between the two models.

\begin{table}[t]
\centering
\small
\setlength{\tabcolsep}{3pt}
\caption{Rationale quality evaluation results (LLM-as-a-Judge) on PuzzleVQA: Qwen3-VL-Flash vs. Gemini 3 Pro.}
\label{tab:tab4}
\begin{tabular}{|l|l|l|l|}
\hline
Criterion & QwenFlash & Gemini 3 Pro & Diff. \\
\hline
Accuracy (\%) & 89.0 & 100.0 & +11.0 \\
Logical Consistency & 50.05 & 89.91 & +39.86 \\
Conciseness and Clarity & 35.75 & 89.90 & +54.15 \\
Reference Alignment & 42.05 & 85.96 & +43.91 \\
\hline
\end{tabular}
\end{table}

\section{Experimental Results}
This section reports results on the unified test set discussed earlier. The analysis follows a sequential workflow: first selecting the strongest baseline, then testing whether additional synthetic sources and data quality interventions can improve it.

\subsection{Initial Baseline Selection}
In the first step, Qwen3VL-8B and Gemma3-12B were trained using synthetic answers generated only by Gemini 3 Pro. The purpose was to identify the most reliable SFT configuration to carry forward. As detailed in Table~\ref{tab6}, where Visual is an alias for Visual Puzzles, Qwen3VL-8B consistently outperformed Gemma3-12B across all dataset mixtures. The best result was obtained by Qwen3VL-8B trained on the PuzzleVQA + MMIQ (3pro) dataset, reaching a general success rate of 48.75\%. Therefore, this specific setting was selected as the new baseline for iterative improvement. Ultimately, the outcomes presented in Table~\ref{tab6} determined the next step: keeping the best baseline fixed and testing whether broader synthetic supervision could further improve generalization.

\begin{table*}[t]
\centering
\caption{Initial Baseline Selection: Model and Data Mixture Comparison on the Test Set}
\label{tab6}
\resizebox{\textwidth}{!}{
\begin{tabular}{|l|l|l|l|l|l|}
\hline
Model & Dataset Configuration (SFT) & General Success (\%) & PuzzleVQA (\%) & BİLSEM (\%) & Visual Puzzle (\%) \\
\hline
Qwen3VL-8B & Visual (3pro) & 32.0 & 34.0 & 42.0 & 20.0 \\
Qwen3VL-8B & MMIQ (3pro) & 28.0 & 36.0 & 24.0 & 24.0 \\
Qwen3VL-8B & PuzzleVQA (3pro) & 36.7 & 52.0 & 24.0 & 34.0 \\
Qwen3VL-8B & Visual + MMIQ (3pro) & 27.3 & 30.0 & 24.0 & 28.0 \\
Qwen3VL-8B & Visual + PuzzleVQA (3pro) & 37.3 & 52.0 & 28.0 & 32.0 \\
Qwen3VL-8B & PuzzleVQA + MMIQ (3pro) & 48.8 & 72.0 & 26.7 & 22.0 \\
Qwen3VL-8B & PuzzleVQA + Visual + MMIQ (3pro) & 32.7 & 38.0 & 32.0 & 28.0 \\
Gemma3-12B & MMIQ (3pro) & 26.0 & 32.0 & 22.0 & 24.0 \\
Gemma3-12B & PuzzleVQA (3pro) & 28.0 & 38.0 & 22.0 & 24.0 \\
Gemma3-12B & Visual + MMIQ (3pro) & 28.0 & 32.0 & 34.0 & 20.0 \\
Gemma3-12B & Visual + PuzzleVQA (3pro) & 31.0 & 40.0 & 34.0 & 20.0 \\
Gemma3-12B & PuzzleVQA + MMIQ (3pro) & 24.0 & 44.0 & 18.0 & 12.0 \\
Qwen3VL-8B & None (Baseline) & 24.7 & 34.0 & 24.0 & 16.0 \\
Gemma3-12B & None (Baseline) & 26.0 & 28.0 & 28.0 & 22.0 \\
Gemini 2.5 Pro & None (Baseline) & 64.0 & 84.0 & 64.0 & 46.0 \\
GPT-5.2 & None (Baseline) & 51.0 & 66.0 & 60.0 & 28.0 \\
\hline
\end{tabular}
}
\end{table*}

\subsection{Baseline Extension and Data-Quality Refinement}
Starting from the selected baseline, and keeping the best dataset mix the same, we first tested whether adding Qwen3VL-Flash synthetic answers improves performance. It did not. Since Qwen3VL-Flash outputs can contain unfinished solutions and incorrect answers, we next applied two data-quality strategies. First, we utilized regex filtering to strictly remove samples with incorrect final answers. Additionally, we employed MultiSFT, a strategy that first regenerates two extra solution traces for the same problem to increase reasoning diversity, and then applies regex filtering to effectively remove any invalid samples.

Table~\ref{tab9} summarizes this second step and reports the best result for each training recipe. Overall, these augmentation strategies improved the standard mixed-data training behavior on the chosen dataset mix, with MultiSFT proving to be the strongest augmentation method. However, the initially selected baseline of standard SFT using only Gemini 3 Pro data remained the strongest configuration overall at 48.75\%. Adding Qwen3VL-Flash data under standard SFT actually degraded performance, dropping general success to 32.25\% for the combined dataset (3pro + QwenFlash) and 30.50\% for QwenFlash alone. Data augmentation partially recovered this degradation, as MultiSFT consistently outperformed regex-only preprocessing in both settings. While MultiSFT delivered the best augmented results achieving 40.50\% for the combined dataset and 32.00\% for QwenFlash alone it ultimately did not surpass the selected baseline.

\begin{table*}[t]
\centering
\caption{Baseline Extension and Data-Quality Refinement on the Test Set}
\label{tab9}
\resizebox{\textwidth}{!}{
\begin{tabular}{|l|l|l|l|l|l|l|}
\hline
Model & Training Data & Preprocessing & General Success (\%) & PuzzleVQA (\%) & BİLSEM (\%) & Visual Puzzle (\%) \\
\hline
Qwen3VL-8B & SFT (3pro) & None & 48.8 & 72.0 & 26.7 & 22.0 \\
Qwen3VL-8B & SFT (3pro + QwenFlash) & None & 32.3 & 41.0 & 23.4 & 24.0 \\
Qwen3VL-8B & SFT (QwenFlash) & None & 30.5 & 38.5 & 23.3 & 20.0 \\
Qwen3VL-8B & SFT (3pro + QwenFlash) & Regex Filter & 27.0 & 36.0 & 20.0 & 12.0 \\
Qwen3VL-8B & SFT (QwenFlash) & Regex Filter & 30.0 & 28.0 & 34.0 & 18.0 \\
Qwen3VL-8B & SFT (3pro + QwenFlash) & MultiSFT & 40.5 & 54.0 & 26.0 & 30.0 \\
Qwen3VL-8B & SFT (QwenFlash) & MultiSFT & 32.0 & 43.0 & 18.0 & 22.0 \\
Qwen3VL-8B & None (Baseline) & None & 24.7 & 34.0 & 24.0 & 16.0 \\
Gemini 2.5 Pro & None (Baseline) & None & 64.0 & 84.0 & 64.0 & 46.0 \\
GPT-5.2 & None (Baseline) & None & 51.0 & 66.0 & 60.0 & 28.0 \\
\hline
\end{tabular}
}
\end{table*}

\subsection{Reasoning Style and Output Efficiency}
A clear pattern was observed in the generated outputs. Better models usually gave direct and compact answers. On the other hand, weaker fine-tuned models often produced long but low-information explanations. These outputs repeated the same idea many times, tried many weak hypotheses, and sometimes failed to give a final parseable answer.

The results suggest that some SFT settings changed the generation behavior of the model. The model started to imitate unstable reasoning traces and spent many tokens on repeated reasoning instead of reaching a clean conclusion, as shown in Table~\ref{tab:model-collapse-stats}.

Table~\ref{tab:model-collapse-example} shows one representative collapse example. The model starts with a reasonable visual observation, but then repeats the same partial rule again and again. 

\begin{table*}[t]
\caption{A clear instance of model collapse observed after SFT of the Qwen3VL-8B model using the PuzzleVQA dataset. The model fails to converge and falls into an infinite repetition loop.}
\label{tab:model-collapse-example}
\centering
\begin{tabular}{|>{\centering\arraybackslash}p{4.5cm}|>{\raggedright\arraybackslash}p{7cm}|}
\hline
\textbf{Question } & \textbf{Generated Output (Model Collapse)} \\
\hline
\vspace{0pt}
\includegraphics[width=0.9\linewidth]{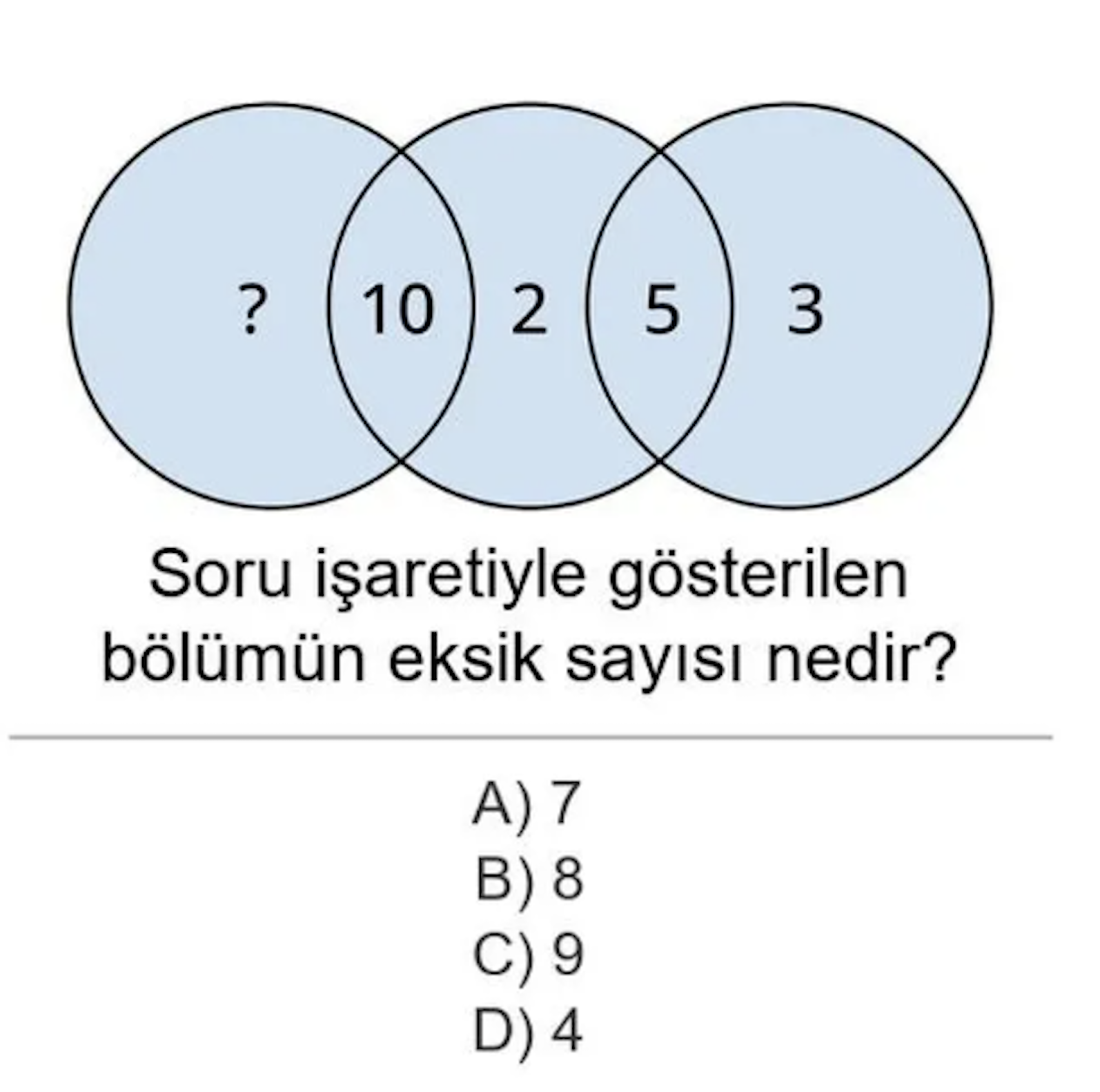}
\newline \vspace{5pt}
\textbf{Correct Answer: B) 8}
& 
\vspace{0pt}
\small
\textit{``Görüntüde örtüşen üç numaralı çember vardır. Birinci çember '?' numaralı, ikinci çember 10 numaralı ve üçüncü çember 3 numaralı... Dolayısıyla desen, örtüşen kısımlardaki sayıların toplamı 10 sayısını verir. Örtüşen kısımlardaki sayıların toplamı 10 olduğuna dayanan desene göre, örtüşen kısımlardaki sayılar toplamı 10 olduğuna göre örtüşen...''} 
\newline\newline
\textbf{[Output truncated due to infinite repetition loop]} \\
\hline
\end{tabular}
\end{table*}

\begin{table*}[t]
\centering
\caption{Output-level diagnostics of SFT-induced collapse on the 400-sample test set.}
\label{tab:model-collapse-stats}
\resizebox{\textwidth}{!}{
\begin{tabular}{|l|l|l|l|l|l|}
\hline
Model & Training Data & General Success (\%) & Empty Output (\%) & Repetition (\%) & Avg. Output Words \\
\hline
Qwen3VL-8B & SFT (3pro) & 48.8 & 1.00 & 9.00 & 215.3 \\
Qwen3VL-8B & SFT (3pro + QwenFlash) & 32.3 & 42.25 & 52.00 & 552.7 \\
Qwen3VL-8B & SFT (QwenFlash) & 30.5 & 42.25 & 51.25 & 553.9 \\
Qwen3VL-8B & MultiSFT (3pro + QwenFlash) & 40.5 & 6.25 & 13.25 & 226.8 \\
\hline
\end{tabular}
}
\end{table*}

Table ~\ref{tab:model-collapse-stats} shows that SFT-induced collapse is a generation-level failure. The SFT (3pro) model gives the best fine-tuned result, with 48.8\% general success and only 1.00\% empty outputs. Its average output length is also short, at 215.3 words. This means that the model usually gives a valid and controlled answer.

However, adding QwenFlash traces strongly changes this behavior. The SFT (3pro + QwenFlash) setting drops to 32.3\% general success. More importantly, it produces empty outputs for 42.25\% of the test samples. The repetition rate also increases to 52.00\%. This means that the model often does not reach a valid answer.

The SFT (QwenFlash) model shows a similar problem. It reaches only 30.5\% general success, has 42.25\% empty outputs, and a 51.25\% repetition rate. These results suggest that the quality of the supervised reasoning traces is very important.

MultiSFT (3pro + QwenFlash) gives a partial recovery. It reduces the empty output rate from 42.25\% to 6.25\%, and the repetition rate from 52.00\% to 13.25\%. It also improves general success to 40.5\%. This shows that using more diverse reasoning paths can make generation more stable.

\clearpage
\section{Conclusion and Limitations}
We empirically demonstrated that Supervised Fine-Tuning (SFT) is a double-edged sword; without high-quality, diverse reasoning traces, SFT can trigger model collapse.

To address the limitations observed in this study, several promising avenues for future research are identified:

\begin{itemize}
    \item \textbf{Diversity in Synthetic Supervision:} Future iterations should involve sampling multiple, distinct reasoning paths from a broader array of models.
    
    \item \textbf{Transitioning from SFT to Reinforcement Learning (RL):} We propose extensive ablation studies on Group Relative Policy Optimization (GRPO). By rewarding the model for the correctness of the final answer and the structural integrity of the rationale, we expect to see a more robust emergence of logical deduction.
    
    \item \textbf{Linguistic and Perceptual Grounding:} To improve performance on Turkish-specific tasks like B\.{I}LSEM, future work should investigate the impact of domain-specific pretraining. 
    
    \item \textbf{Scaling and Benchmarking:} Extending this benchmark to much larger models will help determine if abstract reasoning is primarily an emergent property of scale or if it requires fundamental architectural shifts in how vision and logic are integrated.
\end{itemize}

In summary, while multimodal models can ``see,'' true ``thinking'' requires moving beyond the imitation of patterns toward a verifiable, self-correcting logic.


\section*{Acknowledgements}
This study was supported by the Scientific and Technological Research Council of Turkey (TUBITAK) Grant No: 124E055.
%
%
\bibliographystyle{icml2026}
\bibliography{references}
\end{document}